# Fuzzy Approaches to Abductive Inference

Ndra Mellouli, Bernadette Bouchon-Meunier

Universit Pierre et Marie Curie
LIP6, mailbox 169, 4 place Jussieu
75252 Paris cedex 05, France

**Abstract**: This paper proposes two kinds of fuzzy abductive inference in the framework of fuzzy rule base. The abductive inference processes described here depend on the semantic of the rule. We distinguish two classes of interpretation of a fuzzy rule, certainty generation rules and possible generation rules. In this paper we present the architecture of abductive inference in the first class of interpretation. We give two kinds of problem that we can resolve by using the proposed models of inference.

## Introduction

Abduction, or inference to the best explanation, is a form of inference that starts from data to hypothesis that best explains the data. Charniak and McDermott [1], characterize abduction differently as modus ponens turned backward, inferring the cause of some observed events, generating explanation for what happens as events and generating best explanation. Abduction is an essential component of many tasks as medical diagnosis, scientific discovery, discourse comprehension, etc. It can be represented in the following form proposed by Charles Peirce : "the surprising fact $B$ is observed, but if $A$ were true, $B$ would be a matter of course ; hence there is a reason to suspect that $A$ is true". So it is clear that in abduction, data do not guarantee the truth of hypothesis due to incomplete information. However uncertainty and imprecision play a leading role in human reasoning. The discussion proposed in this paper presents an attempt to formalize abduction by taking into account imprecise and uncertain data in the framework of a fuzzy rule-base. In this context we show that the formalization of abduction is not unique and it depends on the kind of application. This paper is organized as follows. First we define abduction in a fuzzy context, where we need to explore a fuzzy rule base. Next, we try to characterize abduction through two kinds of application which are diagnosis of fault components and causal diagnosis. Finally, future research plans are discussed.

## Abduction in a fuzzy context

For applying fuzzy logic to abduction and particularly to abductive inference, we need to formulate the problem in a general framework.

### Abduction

Abduction is the less well defined type of reasoning. Informal definitions present abduction as "the generation of hypothesis". This definition supposes implicitly a relation is available between hypothesis and observations. This form of abduction can be characterized as "abductive rule inference"[2] and we can represent it as follows:

$$\begin{array}{l}\text{Rule}: A \rightarrow B \\ \text{Observation}: B' \\ \hline \text{Hypothesis}: A' \end{array}$$

The observation $B'$ has a fuzzy characterization which is compared to $B$. For example, we need a rule like " a basket man is tall" and as observation "this man is very tall". Another way to define abduction is to form theories that explain a new observed phenomenon, which is called by "rule abduction"[2] or "abductive rule

generation"[3]. It can be represented as follows:

$$\frac{\text{Observation}: B'}{\begin{array}{l}\text{Rule hypothesis}: A \to B \\ \text{Explanation}: A'\end{array}}$$

In this paper we consider only the first definition of abduction where we suppose that the relation between hypothesis and observation, i.e between $A$ and $B$ is known beforehand. We suppose also that $B'$ is fuzzy, and the corresponding datum is uncertain and imprecise.

## Uncertainty and Imprecision

Uncertainty is inevitable at all levels of human interaction with the environment. At the lowest level, biological processes are for example never clear-cut and without noise. At the cognitive level, uncertainty results from inadequate information sources, limited information processing capacity, or ambiguities in natural language. Another source of uncertainty in data-based systems can be generated from vague concepts due to qualitative reasoning used for example in control or to make a decision. In these contexts the data can be provided in a linguistic way by using a natural language. Some authors [21][22] refer to two kinds of uncertainty: the external and internal one. The external uncertainty occurs by randomness of external events which are something we cannot control. Internal uncertainty refers to ignorance, provided by incomplete knowledge. An important aspect of uncertainty and imprecision management consists in quantifying or measuring them and taking them into account in the abductive process. One possible approach is probability theory, when the only uncertainty is considered. Fuzzy logic is a good candidate to manage both uncertainty and imprecision and to integrate them in the abductive inference.

## Abductive application methodologies and formalization

Representational and inferential capabilities based on imprecise, uncertain, incomplete or inconsistent information are becoming more important in the design, implementation and operation of knowledge-based systems. Knowledge representation has been a major concern of work on knowledge-based systems. The general assumption and then the methodology, has been that, there is domain knowledge which needs to be acquired, quite independently of the problems that we wish to solve. There exist various knowledge formalizations like connectionist networks, logics of various kinds, frame representations and rule-based languages, etc. Each knowledge representation formalism, together with the inference mechanisms using it, defines a unique model. Our model is based on fuzzy rule-based knowledge representation and logical inference, we restrict our attention to "if..then.." rules.

**Definition 1** We note by $\mathcal{R}$: "if $u$ is $A$ then $v$ is $B$" a fuzzy rule such that :

- $A$ and $B$ are two fuzzy characterizations defined respectively on two universes of discourse $U$ and $V$

- $u$ and $v$ are two variables defined respectively on $U$ and $V$

- $A$ and $B$ are characterized by their membership functions $A(u)$ and $B(v)$ defined on the interval $[0, 1]$

■

*Example:* let us consider for example $U$ the set of temperature values in Celcius degrees, $U = [0, 200]$ and $u$ is the temperature variable taking its values on $U$. "Low", "medium" and "high" are fuzzy characterizations on $U$, described by their membership functions. "the temperature is high" is a fuzzy proposition (see FIG.1).

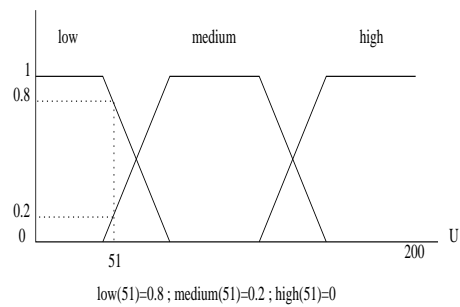

Figure 1: Example of characterizations of the variable "temperature" by means of "low", "medium", "high"

Such fuzzy rule "if $u$ is $A$ then $v$ is $B$" can be interpreted in four ways [4][5] since "$u$ is $A$" and "$v$ is $B$" have two interpretations each. However in the context of knowledge-based systems, the "if" part corresponds to a pattern matching process, which means that the rule is fired when $A(u)$ is high. The "then" part can however be interpreted either as the possibility that $v$ lies in $B$ is high or the certainty that $v$ is characterized by $B$ is high. This gives two kinds of fuzzy "if..then.." rules [4][6]: first, *certainty generation rules* of the form "the higher $A(u)$, the more certain $B(v)$, as "the more crowded the cafeteria, the more certain it is about twelve o'clock", or "the higher $A(u)$, the higher $B(v)$" as "the redder the tomato, the riper it is". Second, we find the *possibility generation rules* of the form "the more $u$ is $A$, the more possible $v$ lies in $B$" (or "the more $u$ is $A$, the larger the set of possible values for $v$ is, around the core [1] of $B$"), as "the more cloudy the sky, the more possible it will rain soon". When an event $B'$ is observed, the mechanism of abductive inference used to construct hypothesis is different in the two cases of interpretation. Before describing the two models of abduction, we define the main properties of the abduction mechanism in a general framework.

**Definition 2** The principal properties of abduction as they are defined by many researchers [7][8][9] are the following:

- the hypothesis $A'$ is consistent with the rule $R$, i.e $\mathcal{R} \cup A'$ is coherent and implies the observation $B'$

- $\mathcal{R} \not\models B'$ and $A' \not\models B'$

- $A'$ is not empty

where $\not\models$ represents a symbol of logical consequence. ∎

The extension of the first property in a fuzzy context is interpreted as $A'$ matches $A$ and $B'$ can be reached by generalized modus ponens from $\mathcal{R}$ and $A'$. The so-called generalized modus ponens, introduced by Zadeh [23], corresponds to the following pattern of inference:

if $u$ is $A$ then $v$ is $B$
$u$ is $A'$
_______________
$v$ is $B'$

where $B'$ is to be computed from $A$, $A'$ and $B$ by equation 1.

**Definition 3** The generalized modus ponens is an extension of the classical modus ponens. The membership function $B'(v)$ of the conclusion is computed as a combination of the membership function $A'(u)$ of the observation and a fuzzy implication $R(A(u), B(v))$ modeling the rule $\mathcal{R}$.

$$\forall v \in V, \quad B'(v) = sup_{u \in U} T(A'(u), R(A(u), B(v))) \quad (1)$$

where $T$ is a generalized modus ponens operator, such that we obtain $B' = B$ when $A'=A$ (in this case, the generalized modus ponens is coherent with the classical modus ponens) [23]. ∎

In a relational context, equation (1) can be interpreted as a direct image [10] of a fuzzy set $A'$ under a fuzzy relation $R$ if we consider that $R$ is a fuzzy relation from $U$ to $V$. A fuzzy implication $R$ models the rule $\mathcal{R}$ by quantifying the relation degree between premise "$u$ is $A$" and conclusion "$v$ is $B$". We distinguish three classes of fuzzy implications:

1. *s-implication*[2]:
   $R(A(u), B(v)) = C(\overline{A}(u), B(v))$ where $C$ is a t-conorm and $\overline{A}$ is the complement of $A$.

2. *r-implication*[3]:
   $R(A(u), B(v)) = sup\{z \in [0,1], T(A(u), z) \leq B(v)\}$ for a t-norm $T$

3. *Ql-implication*[4]:
   $R(A(u), B(v)) = \overline{T(A(u), \overline{T(A(u), B(v))})}$

In the rest of this paper we define abduction in the framework of certainty generation rules, the case of possibility generation rules suggests further studies.

## Abduction and certainty generation rules

Fuzzy rule-based systems have been mainly used as a convenient tool to handle control laws from data. Recently, in a knowledge representation-oriented perspective [11], a typology of fuzzy rules has been laid bare, by emphasizing the distinction between certainty generation rules and variation generation ones. There are two main kinds of certainty generation rules which are *certainty rules* and *variation rules*.

---

[1] Core of $B$ is a set of elements $v$ in V such that $B(v) = 1$

[2] We start from the classical logic form $A \rightarrow B = \neg A \vee B$ and we extend it to fuzzy logic by substituting the negation by a fuzzy negation, the disjonction by a fuzzy disjonction.

[3] We start from the classical logic form $A \wedge (\neg A \vee B) = A \wedge B$, we remark that the truth value of $A \wedge (\neg A \vee B)$ is not greates than the truth value of $B$, and then we substitute logical operators by fuzzy operators.

[4] We start from the classical logic form $\neg A \vee (A \wedge B) = \neg(A \wedge \neg(A \wedge B))$ and we substitute logical operators by fuzzy operators.

## Certainty rules and inference

Certainty rules are of the form "the more $u$ is $A$ the more certainly $v$ lies in $B$", for example "the younger a man, the more certainly he is single" (according to this example, crisp consequences which are particular cases of fuzzy consequences can be taken into account by our study). These rules are modeled with fuzzy implications of the first class. They consist in a simple extension of the binary implication $\overline{A} \vee B$. Then if we consider an observation $B'$ that we represent by the proposition "$v$ is $B'$", we look for the characterization $A'$ which satisfies the main properties of abduction given by definition 1 and equation (1). We note $S$ a fuzzy s-implication. Equation (1) becomes:

$\forall v \in V,$
$$B'(v) = sup_{u \in U} T(A'(u), S(A(u), B(v))) \quad (2)$$

The abductive problem consists in computing the antecedents of $B'$ under the fuzzy relation given by the implication $S$ where $A'$ is the unknown element. Equation (2) has a non empty solution if the necessary (and not sufficient) condition $\forall v \in V, B'(v) \leq sup_{u \in U} A'(u)$ [10] is satisfied. Let us suppose that equation (2) has a non empty solution, then to solve such equation, we need to characterize the following reasoning:

Rule : $\overline{B} \to \overline{A}$

Observation : $B'$

─────────────

Hypothesis : $A'$

It can be proven that any solution $A'$ of (2) is given by the following equation: $\forall u \in U,$

$$A'(u) = sup_{v \in V} T(B'(v), S(\overline{B}(v), \overline{A}(u))) \quad (3)$$

Moreover equation (3) implies an equivalence between the two rules " if $u$ is $A$ then $v$ is $B$" and "if $v$ is $\overline{B}$ then $u$ is $\overline{A}$", and $S$ must have a contrapositive symmetry property.[5] It is the case for any s-implication. Table

---
[5]Contrapositive symmetry:

$S(A(u), B(v)) = C(\overline{A}(u), B(v))$ where $C$ is a t-conorm
$= S(\overline{B}(v), \overline{A}(u))$

| s-implications: $S(A(u), B(v))$ |
|---|
| $1 - A(u) + A(u).B(v)$ |
| $max(1 - A(u), min(A(u), B(v)))$ |
| $max(1 - A(u), B(v))$ |
| $min(1, B(v) - A(u) + 1)$ |

Table 1: Examples of s-implications

(TAB.1) gives examples of well-known s-implications.

## Variation rules and inference

Variation rules are of the form "the more $u$ is $A$ the more $v$ is $B$ as in "the redder the tomato, the riper it is". These rules are modeled by r-implications. They have nice and suitable properties which allow us to characterize the solution $A'$ and to get easily an expression of the image equation of $B'$ deduced from equation (1). A few properties of such fuzzy implication operators are given below [13]:

**Property 1:**
A binary implication operator $I_T : [0,1]^2 \to [0,1]$ is associated to a t-norm $T$ such that:
$I_T(a,b) = sup\{z \in [0,1], T(a,z) \leq b\}$ and it satisfies the following properties, $\forall (a,b,c) \in [0,1]^3$:

1. if $b \leq c$ then $I_T(a,b) \leq I_T(a,c)$
2. $T(a, I_T(a,b)) \leq b$
3. $I_T(a, T(a,b)) \geq b$
4. if $a \leq b$ then $I_T(a,c) \geq I_T(b,c)$
5. if $a \leq b$ then $I_T(a,b) = 1$
6. $I_T(1,b) = b$
7. $b \leq I_T(a,b)$

These operators are much studied by many researchers. The dependency between the choice of the t-norm and the form of the implication was given by several studies in the literature [12][13].

The form of the equation (1) after taking into account the implication $I_T$ is the following : $\forall v \in V,$

$$B'(v) = sup_{u \in U} T(A'(u), I_T(A(u), B(v))) \quad (4)$$

The abductive problem we define in the framework of the variation rules, consists in computing the antecedents $A'$ of $B'$ under the fuzzy relation given here by the implication $I_T$. We suppose that the equation (4) has a non empty solution for the same condition

| r-implications: $I_T(A(u), B(v))$ |
|---|
| 1 if $A(u) \leq B(v)$ and $B(v)$ otherwise |
| $min(\frac{B(v)}{A(u)}, 1)$ if $A(u) \neq 0$ |
| and 1 otherwise |
| $min(1, B(v) - A(u) + 1)$ |

Table 2: Examples of r-implications

as certainty rules, we prove that any solution $A'$ antecedent of $B'$ is given by equation (5):

**Proposition 1** Let A and $A'$, B and $B'$ four fuzzy sets respectively on U and V, such that $B'$ is obtained from A, $A'$, B by means of equation (4). Then: $\forall u \in U$,

$$A'(u) \leq inf_{v \in V} I_T(I_T(A(u), B(v)), B'(v)) \quad (5)$$

**Proof**
if $\forall v \in V$
$B'(v) = sup_{u \in U} T(A'(u), I_T(A(u), B(v)))$
then $\forall u \in U$ $T(A'(u), I_T(A(u), B(v))) \leq B'(v)$.
Since $T$ is commutative, we have
$T(I_T(A(u), B(v)), A'(u)) \leq B'(v)$.
If $I_T$ satisfies inequation 1 of property 1
it holds that $\forall u, v \in U \times V$ :
$I_T(I_T(A(u), B(v)), T(I_T(A(u), B(v)), A'(u))$
$\leq I_T(I_T(A(u), B(v)), B'(v))$.
If $I_T$ verifies inequation 3 of property 1
we have :
$A'(u) \leq$
$I_T(I_T(A(u), B(v)), T(I_T(A(u), B(v)), A'(u))$
Since
$I_T(I_T(A(u), B(v)), T(I_T(A(u), B(v)), A'(u))$
$\leq I_T(I_T(A(u), B(v)), B'(v)), \forall u, v \in U \times V$
We get $A'(u) \leq I_T(I_T(A(u), B(v)), B'(v))$.
And $inf_{v \in V} I_T(I_T(A(u), B(v)), B'(v)) \geq A'(u)$.

The equality is reached for instance when $B'=B$ and $I_T$ is Gdel implication. In this case we get the Bandler-Kohout superdirect image [10] of B under $I_T$. Moreover, the problem of abduction is reduced in some particular cases to solving a fuzzy arithmetic equation[14][15][16]. Table (TAB.2) gives an overview of well-known r-implications.

## Example of applications

There has been an explosion of recent work in artificial intelligence that recognizes the importance of abductive reasoning. Many formal researches [17][18][19] [20] and applications including medical diagnosis, fault diagnosis, natural language understanding, scientific discovery have been characterized as abduction. In this Section we try to underline the connection between a kind of application and a type of abductive inference model or architecture. We consider here two kinds of abductive problem diagnosis, diagnosis of fault components and causal diagnosis. For each of these problems, we choose a convenient model of abductive inference which is coherent with the nature of data.

### Diagnosis of fault components

The diagnosis of fault components is much common in analogic circuit (AC) systems. Despite the big number of research in such domains the results are little satisfactory because of several difficulty factors. For example a continuous analogic signal causes uncertainty and imprecision to evaluate some electronic parameters like the intensity of the current power. In practice, the supervision of most complex AC systems is automated. Most of their correct behavior is represented by a rule base. The diagnosis of fault component consists in following the origin component such that the contrary of its usual value has occurred. In other words, if we denote by $B'$ the surprising event, there exists a characterization B of some component such that $B'$ matches its contrary value. That is why a formulation of abductive inference using the contrapositive symmetry model is useful to detect and characterize the fault.

### Causal Diagnosis

Causal diagnosis is considered as an abductive problem. Hence, in medical application, abduction is needed to justify the expert answer. However, the answer justification refers to the ability of an expert system to explain how or why he arrives at certain conclusions such as a patient's differential diagnosis. To do that, we suppose that there exists a connection between causes and effects via a causal relation. The most commonly used approach for modeling such relation has been to store medical knowledge as a collection of "if..then" rules and to make inferences about a patient using deductive logic. In contrast, empirical studies suggest that the diagnosis reasoning is better characterized as an abductive inference process. This approach can be captured in a fuzzy framework as a simple scheme: for given linguistic premises $A_i$ (causes), conclusions $B_i$ (effects) and a fuzzy observation $B'$ we try to define abductive explanations represented by the unknown hypothesis $A'$. The model of inference of this case refers to the second model of abductive inference that we have defined in the last section. It allows us to compute the uncertainty and imprecision of each hypothesis calculated from the equation (5) and so the degree of contribution of each value to cause the observation or the effect $B'$.

## Conclusion and future work

The concept of explanation, diagnosis hypothetical reasoning are considered as abductive problems. In Artificial intelligence, the most commonly used approach for modeling knowledge associated to such problems is to store a collection of "if .. then" rules. In a fuzzy framework, the model chosen to represent the abductive inference process neglects the semantic of rules. This paper provides two distinct possible architectures of abductive inference which take into account the meaning of the relation between premises and conclusions of rules. As a first step of our study, we are considered only a single rule with two kinds of semantics and we have presented two examples of application that can be handled. In a future work we propose to give a detailed study on abductive inference in the framework of certainty generation rules and possible generation rules using a set of rules and a set of input and output variables. We will study more particularly how we quantify the uncertainty and imprecision and how to interpret them. These results will be tested on a real-world problem of diagnosis.